\documentclass{article}

    \usepackage[preprint,nonatbib]{neurips_2020}

\usepackage[utf8]{inputenc} %
\usepackage[T1]{fontenc}    %
\usepackage{hyperref}       %
\usepackage{url}            %
\usepackage{booktabs}       %
\usepackage{amsfonts}  
\usepackage{amsmath,amssymb}
\usepackage{nicefrac}       %
\usepackage{microtype}      %

\usepackage{todonotes}
\usepackage{dsfont}
\usepackage{graphicx}
\usepackage{multirow}
\usepackage{fancyvrb}
\usepackage{wrapfig}

\title{Tempered Sigmoid Activations for\\ Deep Learning with Differential Privacy}

\author{Nicolas Papernot, Abhradeep Thakurta, Shuang Song, Steve Chien, \'Ulfar Erlingsson\thanks{\'Ulfar Erlingsson is currently at Apple, work done while the author was at Google Brain.} \\
	Google Brain}
\begin{document}

\maketitle

\begin{abstract}
Because learning sometimes involves sensitive data, machine learning algorithms
have been extended to offer privacy for training data. In practice, this has
been mostly an afterthought, with privacy-preserving models obtained by
re-running training with a different optimizer, but using the model
architectures that already performed well in a non-privacy-preserving setting.
This approach leads to less than ideal privacy/utility tradeoffs, as we show
here. Instead, we propose that model architectures are chosen  \emph{ab initio}
explicitly for privacy-preserving training.

To provide guarantees under the gold standard of differential privacy, one must
bound as strictly as possible how individual training points can possibly affect
model updates. In this paper, we are the first to observe that the choice of
activation function is central to bounding the sensitivity of privacy-preserving
deep learning. We demonstrate analytically and experimentally how a general
family of bounded activation functions, the tempered sigmoids, consistently
outperform unbounded activation functions like ReLU. Using this paradigm, we
achieve new state-of-the-art accuracy on MNIST, FashionMNIST, and CIFAR10
without any modification of the learning procedure fundamentals or differential
privacy analysis.

\end{abstract}

\section{Introduction}

Machine learning (ML) can be usefully applied
to the analysis of sensitive data, 
e.g., in the domain of 
healthcare~\cite{kononenko2001machine}. 
However,
ML models may
unintentionally reveal sensitive aspects
of their training data,
e.g., due to
overfitting~\cite{shokri2017membership,song2019auditing}.
To counter this, 
ML techniques 
that offer strong 
 guarantees expressed in the framework of differential privacy~\cite{dpbook}
have been developed.
A seminal example is the
differentially private stochastic gradient 
descent, or  DP-SGD, of Abadi et al.~\cite{abadi2016deep}.
The technique is a generally-applicable modification
of stochastic gradient descent.
In addition to its rigorous privacy guarantees, it has been empirically shown to stop known attacks against the privacy of training data; a representative example being the
 leaking of secrets~\cite{secretsharer}.

Beyond privacy, training using DP-SGD 
offers
 advantages 
such as strong generalization
and the promise of 
reusable holdouts~\cite{tfprivacy, reusable}.
Yet,
its advantages have not been without cost:
empirically, the test accuracy of
differentially private ML
 is 
consistently lower than that
of non-private learning (e.g., see ~\cite{scalable}).
Such accuracy loss may sometimes be
inevitable: for example, the task
may involve heavy-tailed distributions
and noise added by DP-SGD  
hinders visibility of examples in the tails~\cite{feldman2019does,shmatikovimpact}.
However,
this does not explain the accuracy loss of
differentially private ML on
 benchmarks
that are known to be relatively simple when learning without privacy: e.g., 
MNIST~\cite{yann1998mnist}, 
FashionMNIST~\cite{xiao2017/online}, and
CIFAR10~\cite{krizhevsky2009learning}.

An important step in providing differential privacy guarantees
for an algorithm is to assess its \textit{sensitivity}.
A learning algorithm's sensitivity characterizes how much an individual
training point can, in the worst case, affect the learning algorithm's outputs
(i.e., values of the model parameters).
The ability to more strictly bound sensitivity 
leads to stronger privacy guarantees. 
To strictly bound the impact of 
any training example,
DP-SGD makes two changes to every step of gradient-descent optimization:
first,  each example's
gradient contribution is limited to a fixed bound
(in practice, by clipping all per-example
gradients to a maximum $\ell_2$ norm); 
second, 
random (Gaussian) noise 
of the scale of the clipping norm
is added to each batch's 
combined gradient,
before it is backpropagated to update
model parameters.
Together, 
these changes 
create a new, artificial noise floor at each
step of gradient descent,
such that the unique signal of
any individual
example is 
below this new noise floor;
this allows
differential privacy
to be guaranteed for all  
training examples~\cite{dpbook}.

This paper is the first to observe that DP-SGD leads to exploding model activations as a deep neural network's training progresses. This makes it difficult to control the training algorithm's sensitivity at a minimal impact to its correctness. Indeed, exploding activations cause unclipped gradient magnitudes to also increase, which in turn induces an information loss once the clipping operation is applied to bound gradient magnitudes. This exacerbates the negative impact of noise calibrated to the clipping bound, thus degrading the utility of each gradient step
when learning with privacy. Indeed, the gradient clipping of DP-SGD does not bring the nice properties of gradient clipping commonly used to regularize deep learning~\cite{zhang2019gradient} because DP-SGD clips gradients at the granularity of individual training examples rather than at the level of a batch.

We thus hypothesize that activation functions need to be bounded when learning with DP-SGD. We propose that neural architectures for private learning employ a general family of bounded activations: \textit{tempered sigmoids}. We note that prior work has explored tempered losses as a means to provide robustness to noise during training~\cite{amid2019robust}. Because the family of tempered sigmoids can---in the limit---represent an approximation of  ReLUs~\cite{nair2010rectified} on the subset of their domain that is exercised in training, we expect that our approach will perform no worse than current architectures. These architectures use ReLUs as the de facto choice of activation function. 

Through both analysis and experiments, we validate the significantly superior performance of tempered sigmoids when training neural networks with DP-SGD.
In our analysis, we relate the role of the  temperature parameter in tempered sigmoids to the clipping operation of DP-SGD. Unlike prior work, which attempted to adapt the clipping norm to the gradients of each layer's parameters post hoc to training~\cite{mcmahan2018general}, we find that tempered sigmoids preserve more of the  signal contained in gradients of each layer because they rescale each layer's activations and better predispose the corresponding layer's gradients to clipping. 
 We conclude that using tempered sigmoids is a better default activation function choice for private ML.
 
In summary, our contributions 
 facilitate  DP-SGD learning 
as follows:

\begin{itemize}
\item We analytically show in Section~\ref{sec:approach} how tempered sigmoid activations control the gradient norm explicitly, and in turn support faster convergence in the settings of differentially private ML, by eliminating the
negative effects of 
clipping and noising large gradients.
\item To demonstrate empirically the superior performance of tempered sigmoids, we show in Section~\ref{sec:family} how using tempered sigmoids instead of ReLU activations
significantly improves a model's private-learning 
suitability and achievable
privacy/accuracy tradeoffs. 
\item We advance  the state-of-the-art of deep learning with differential privacy for
MNIST, FashionMNIST, and CIFAR10. On these datasets, we find  in Section~\ref{sec:tanh} that the parameter setting in which tempered sigmoids perform best happens to correspond to the \texttt{tanh} function. On MNIST, our model achieves 98.1\% test accuracy for a privacy guarantee of $(\varepsilon, \delta)=(2.93, 10^{-5})$, whereas the previous state-of-the-art reported in the TensorFlow Privacy library~\cite{tfprivacy} was 96.6\%. On FashionMNIST, we obtain $86.1\%$ test accuracy compared to $81.9\%$ for $(\varepsilon, \delta)=(2.7, 10^{-5})$. Finally, on CIFAR10, we achieve 66.2\% test accuracy at $(\varepsilon, \delta)=(7.53, 10^{-5})$ in a setup for which prior work achieved 61.6\%.
\end{itemize}
\section{Training-data Memorization, 
Differential Privacy, and DP-SGD}
Machine learning models easily memorize 
sensitive, personal, or private data that
was used in their training, and models
may in practice disclose this data---as
 demonstrated by membership inference attacks~\cite{shokri2017membership} and secret extraction results~\cite{song2019auditing,secretsharer}.
 
To reason about the privacy guarantees 
of algorithms such as training by stochastic gradient descent,
differential privacy has become
the established gold standard~\cite{dpbook}.
Informally, an algorithm is differentially private if it  always produces effectively the same output (in a mathematically precise sense), when applied to two input datasets that differ by only one record.
Formally, a learning algorithm $A$ that trains 
models from the set $S$ is $(\varepsilon,\delta)$-differentially-private, if the following holds for all training datasets $d$ and $d'$ that differ by exactly one record:
\begin{equation}
    \Pr[A(d)\in S] \leq e^\varepsilon \Pr[A(d')\in S] + \delta
\end{equation}
Here, $\varepsilon$ gives the formal privacy guarantee, by placing a strong upper bound on any privacy loss, even in the worst possible case.  A lower $\varepsilon$ indicates a stronger privacy guarantee or a tighter upper bound.
The factor $\delta$ allows for some probability that the property may not hold (in practice, this $\delta$ is required to be very small, e.g., in inverse proportion to the dataset size).

A very attractive property of differential-privacy guarantees is that they hold true for all attackers---whatever they are probing and whatever their prior knowledge---and that they remain true under various forms of composition.
In particular, 
the output of a differentially-private algorithm can be arbitrarily post processed, without any weakening of the guarantees.
Also, if sensitive training data contains 
multiple examples from the same person (or, more generally, the same sensitive group),
$\varepsilon$-differentially-private training 
on this data will result in model with a
$k\varepsilon$-differential-privacy guarantee for each person,
as long as at most $k$ training-data records
are present per person.

Abadi et al.~\cite{abadi2016deep} introduced DP-SGD as a method for training deep neural networks 
with differential-privacy guarantees that was able to achieve better privacy and utility
than previous efforts~\cite{chaudhuri2011differentially,song2013stochastic,bassily2014private}.
DP-SGD bounds 
the sensitivity of the learning process to each individual training example
by computing  per-example gradients $\{g_i\}_{i\in 0..n-1}$ with respect to the loss,
for the $n$ model parameters $\{\theta_i\}_{i\in 0..n-1}$, 
and clipping each per-example gradient to a maximum fixed $\ell_2$ norm $C$:
Subsequently, to the average of these per-example gradients,
DP-SGD adds (Gaussian) noise  whose standard deviation $\sigma$ is proportional to this sensitivity.
In this work, we use the canonical implementation of DP-SGD and its associated analysis from the TensorFlow Privacy library~\cite{tfprivacy}.
\section{Approach}
\label{sec:approach}

When training a model with differential privacy, gradients computed during SGD are computed individually for each example (i.e., the gradient computation is not averaged across all samples contained in a minibatch). The gradient $g_i$  for each model parameter $\theta_i$ is then clipped such that the total $l_2$ norm of the gradient across all parameters is bounded by $C$:
\begin{equation}
\label{eq:clipping}
g_i \leftarrow g_i \cdot \min\left(1, \frac{C}{\sqrt{\sum_{i=0}^{n-1}{g_i^2}}}\right)
\end{equation}
Because this operation is performed on per-example gradients, this allows DP-SGD to control the sensitivity of learning to individual training examples. However, this clipping operation will lead to information loss when some of the signal contained in gradients is discarded because the magnitude of gradients is too large. One way to reduce the magnitude (or at least control it), is to prevent the model's activations from exploding. This is one of the reasons why  common design choices for the architecture of modern deep neural networks make it difficult to optimize model parameters with DP-SGD: prominent activation functions like the REctified Linear Unit (ReLU) are unbounded. 

We hypothesize that replacing ReLUs with a bounded activation function prevents activations from exploding and thus keeps the magnitude of gradients to a more reasonable value. This in turn implies that, given a fixed level of privacy guarantee, the clipping operation applied by DP-SGD will discard less signal from gradient updates---eventually resulting in higher performance at test time.

\paragraph{Tempered sigmoids.} Based on this intuition, we propose replacing the unbounded activations typically used in deep neural networks with a general family of bounded activations: the tempered sigmoids. We note that an idea that is conceptually close to ours, the use of tempered losses, was recently found to provide robustness to noise during training~\cite{amid2019robust}.\footnote{We experimented with the tempered loss of~\cite{amid2019robust} but did not find any improvements for DP-SGD training.} Tempered sigmoids are the family of functions that take the form of: 
\begin{equation}
     \phi_{s,T,o}: x\mapsto \frac{s}{1+e^{- T\cdot x}} - o
     \label{eq:ad1}
\end{equation}
where $s$ controls the scale of the activation,  $T$ is the inverse temperature, and $o$ is the offset. By decreasing the value of $s$, we reduce the magnitude of a neuron's activation. Complementary to this, the inverse temperature rescales a neuron's weighted inputs.  We note that setting $s=2$, $T=2$, and $o=1$ in particular yields the tanh function exactly, i.e., we have $\phi_{2,2,1}=\texttt{tanh}$. 

\begin{figure}[h]
\centering
\includegraphics[width=6.5cm]{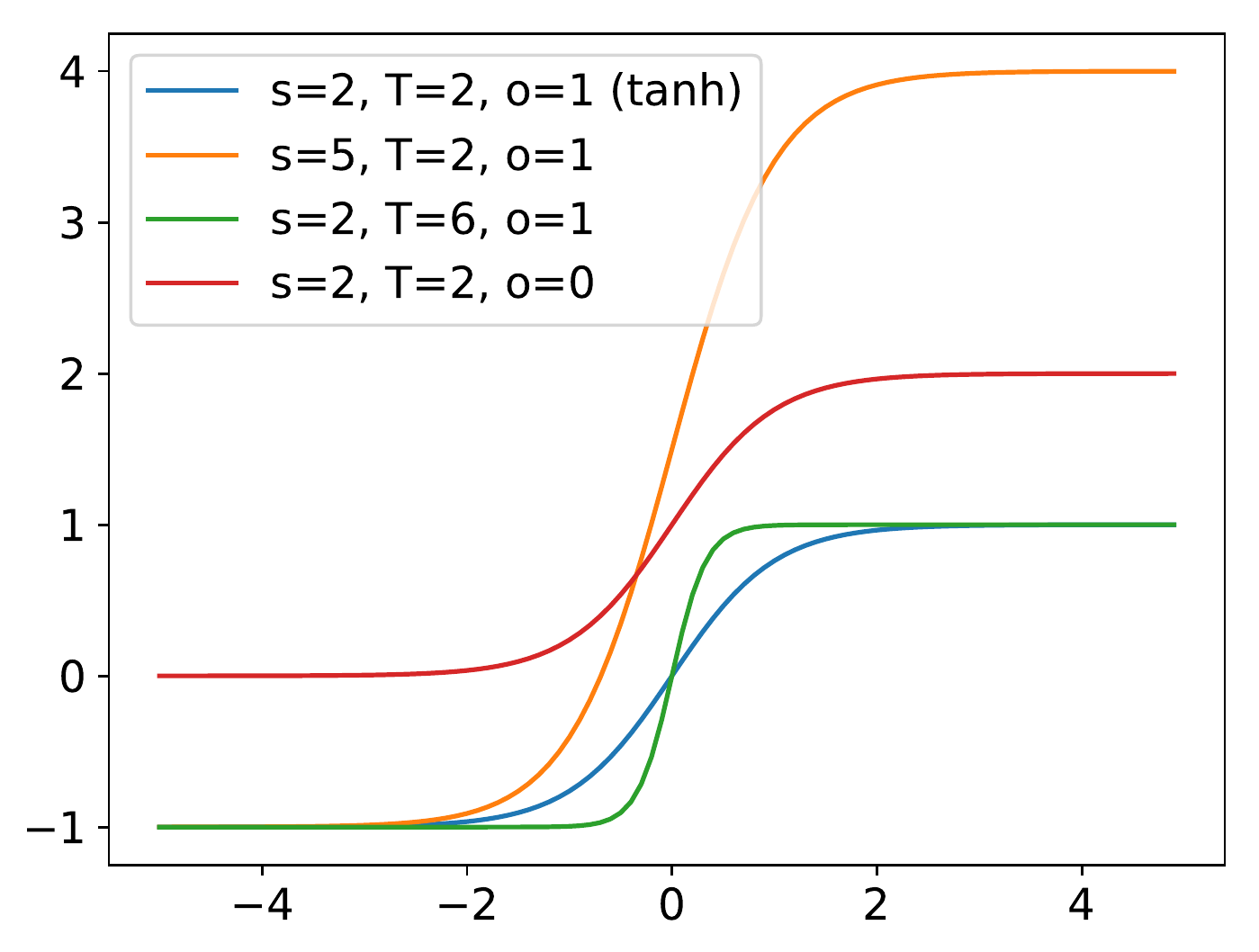}
\caption{Tempered sigmoids: we plot examples for representative values of the scale $s$, inverse temperature $T$, and offset $o$. The blue line corresponds to the parameter triplet $(s=2,T=2,o=1)$ where the tempered sigmoid is exactly a \texttt{tanh}.}
\label{fig:tempered-sigmoids}
\end{figure}

\paragraph{Controlling the gradient norm with tempered sigmoids.} One of the main issues in practice with DP-SGD is tuning the value of the clipping parameter:
\begin{itemize}
    \item If $C$ is set too low, then clipping introduces bias by changing the underlying objective optimized during learning. 
    \item Instead if the clipping parameter $C$ is set too high, clipping increases variance by forcing DP-SGD to add too much noise. Indeed, recall that DP-SGD adds Gaussian noise with  variance $\sigma^2$ to the average of (clipped) per-example gradients. This noise is scaled to the clipping norm such that $\sigma^2=M^2 C^2$ where $M$ is a hyperparameter  called the noise multiplier. Thus, large clipping norms lead to noise with large variance being added to the average gradient before it is applied to update model parameters.
\end{itemize}
It turns out that the temperature parameter $T$ in Equation~\eqref{eq:ad1} can be used as a knob to control the norm of the gradient of the loss function, and with an appropriate choice avoids these two issues of clipping.  In the following, we formalize the relationship between our tempered sigmoids and the clipping of DP-SGD in the context of the binary logistic loss and its multiclass counterpart. 

Consider the tempered logistic loss: $\ell(\theta;z,y)=\ln\left(1+\exp(-y\cdot T\cdot\langle z,\theta\rangle)\right)$, where $z,\theta\in\mathbb{R}^d$, $y\in\{-1,+1\}$, and $T\in\mathbb{R}$ is the \emph{inverse} temperature. Notice that the above expression is an instantiation of $\ln\left(1/\phi_{s,T,o}\right)$, where $s=1$ and $x=y\cdot\langle z,\theta\rangle$. Now, if we take the gradient of $\ell$ w.r.t $\theta$, we have the following for the $\ell_2$-norm of the gradient.
\begin{equation}
    \|\nabla_\theta \ell\|_2 =\left\|\frac{-T\cdot y\cdot z}{1+\exp(T\cdot y\cdot\langle z,\theta\rangle )}\right\|_2\leq |T|\cdot \|z\|_2
    \label{eq:1232}
\end{equation}
We observe the following two things from~\eqref{eq:1232}:
i) Controlling $T$ directly controls the norm of the gradient of $\ell$, and hence controls clipping norm in general (when using tempered sigmoid as an activation), ii) Specifically, for logistic loss, $T$ can be thought of as linear scaling of the feature vector $z$, when the point $(z,y)$ is ``grossly misclassified''. These observations suggest that one can use the inverse temperature to control the norm of the gradient, and may not ever cross the ``clipping threshold'' in DP-SGD. 

One can extend this observations to multiclass logistic loss $\ell(\theta;z,y)=\ln\left(\frac{\exp(T\cdot\langle z,\theta y\rangle)}{\sum_{j\in 1.. k}\exp(T\cdot\langle z,\theta_j\rangle)}\right)$ where $y$ is now a one-hot label vector and $\theta\in \mathbb{R}^{d\times k}$ for a problem with $k$ classes. The partial gradient of $\ell$ w.r.t. $\theta_m$, for any class $m\in 1..k$, becomes: 
\begin{equation}
    \partial_{\theta_m} \ell = \left( \mathds{1}_{m=y} - \frac{\exp(T\cdot\langle z,\theta_m\rangle)}{\sum_{j\in 1..k} \exp(T\cdot\langle z,\theta_j\rangle)} \right) \cdot T \cdot z
\end{equation}
Because the norm of the expression in parenthesis is smaller than 1, we thus have that $\| \partial_{\theta_m} \ell \|_2 \leq |T| \cdot \|x\|_2$. From this we derive that the norm of the gradient $\nabla_\theta \ell = [\partial_{\theta_1} \dots \ \partial_{\theta_k}]$ would correspondingly be bound by $\sqrt{k}\cdot \|\partial_{\theta_m} \ell\|_2 \leq \sqrt{k}\cdot|T|\cdot\|z\|_2$, where $k$ is the number of classes.

\section{Experimental Setup}

We use three common benchmarks for differentially private ML: MNIST, FashionMNIST, and CIFAR10. While the three datasets are considered as ``solved'' in the computer vision community, achieving high utility with strong privacy guarantees remains difficult on all three  datasets~\cite{abadi2016deep,tfprivacy}. Concretely, the state-of-the-art for MNIST is a test accuracy of $96.6\%$ given an $(\varepsilon, \delta)=(2.93, 10^{-5})$ differential privacy guarantee. With stronger guarantees, the accuracy continues to degrade. In the same privacy-preserving settings, prior approaches achieve a test accuracy of $81.9\%$ on FashionMNIST. For CIFAR10, a test accuracy of $61.6\%$  can be achieved given an $(\varepsilon, \delta)=(7.53, 10^{-5})$ differential privacy guarantee.

All of our experiments are performed with the JAX framework in Python, on a machine equipped with a 5th generation Intel Xeon processor and NVIDIA V100 GPU acceleration. For both MNIST and FaashionMNIST, we use a convolutional neural network whose architecture is described in Table~\ref{tbl:architecture}. For CIFAR10, we use the deeper model in Table~\ref{table:cifar10_allconv}. The choice of architectures is motivated by prior work which showed that training larger architectures is detrimental to generalization when learning with privacy~\cite{bassily2014private}. This can be explained in two ways. Given a fixed privacy guarantee, increasing the number of parameters increases (a) how much each parameter needs to be clipped relatively and (b) how much noise needs to be added, with the  norm of noise increasing as a function of the square root of the number of parameters. 

When we train these architectures with ReLU activations for both the convolution and fully-connected layers, we are able to exactly reproduce the previous state-of-the-art results mentioned above for MNIST, FashionMNIST, and CIFAR10.
To experiment with the tempered sigmoid proposed in Section~\ref{sec:approach}, we implement it in JAX and use it in lieu of the ReLU in the architecture from Table~\ref{tbl:architecture} and Table~\ref{table:cifar10_allconv}. Our code is staged for an open-source release, and we include the code snippet for the tempered sigmoid activation below---to demonstrate the practicality of implementing the change we propose in neural architectures.

\begin{Verbatim}[fontsize=\small]
from jax.scipy.special import expit

def tempered_sigmoid(x, scale=2., inverse_temp=2., offset=1., axis=-1):
    return scale * expit(inverse_temp * x) - offset

def elementwise(fun, **fun_kwargs):
    """Layer that applies a scalar function elementwise on its inputs."""
    init_fun = lambda rng, input_shape: (input_shape, ())
    apply_fun = lambda params, inputs, **kwargs: fun(inputs, **fun_kwargs)
    return init_fun, apply_fun

TemperedSigmoid = elementwise(tempered_sigmoid, axis=-1)
\end{Verbatim}

\begin{table}[b]
\parbox{.45\linewidth}{
\centering

    \begin{tabular}{|c|c|}
    \multicolumn{1}{c}{Layer} & 
    \multicolumn{1}{c}{Parameters}  \\
    \hline
    \hline
    Convolution & $16$ filters of 8x8, strides 2 \\
    Max-Pooling & 2x2 \\
    Convolution & $32$ filters of 4x4, strides 2  \\
    Max-Pooling & 2x2 \\
    Fully connected & 32 units  \\
    Softmax & 10 units \\
    \hline
    \end{tabular}
    \caption{Convolutional model architecture.}
    \label{tbl:architecture}

}
\hfill
\parbox{.5\linewidth}{
\centering

\begin{tabular}{ |c|c| } 
\multicolumn{1}{c}{Layer} & 
\multicolumn{1}{c}{Parameters}  \\
\hline \hline
Convolution  $\times 2$ & 32 filters of $3 \times 3$, strides 1\\
Avg-Pooling & $2 \times 2$, stride 2 \\
Convolution  $\times 2$ & 64 filters of $3 \times 3$, strides 1 \\
Avg-Pooling & $2 \times 2$, stride 2 \\
Convolution  $\times 2$ & 128 filters of $3 \times 3$, strides 1 \\
Avg-Pooling & $2 \times 2$, stride 2 \\
Convolution & 256 filters of $3 \times 3$, strides 1 \\
Convolution & 10 filters of $3 \times 3$, strides 1 \\
Averaging & over spatial dimensions \\
\hline
\end{tabular}
\caption{CIFAR10 model architecture.}
\label{table:cifar10_allconv}

}
\end{table}

\section{Evaluating the family of tempered activation functions}

\subsection{Improved privacy-utility tradeoffs with tempered sigmoids}
\label{sec:family}

For each of the three datasets considered, we use DP-SGD to train a pair of models. The first model uses ReLU whereas the second model uses a tempered sigmoid $\phi_{s,T,o}$ as the activation for all of its hidden layers (i.e., both convolutional and fully-connected layers). The models are based off the architecture of Table~\ref{tbl:architecture} for MNIST and FashionMNIST, or Table~\ref{table:cifar10_allconv} for CIFAR10.  All other architectural elements are kept identical. In our experiments,
we subsequently fine-tuned architectural aspects (i.e., model capacity) as well as the choice of optimizer and its associated hyperparameters,
separately
for the activation function in each setting (ReLU and tempered sigmoid),
to avoid favoring any one choice.

Recall from Section~\ref{sec:approach} that tempered sigmoids $\phi_{s,T,o}$ are bounded activations that are parameterized such that their inputs and output can be rescaled---through the inverse temperature $T$ and scale $s$ parameters respectively---and their output recentered with the offset $o$.
Tempered sigmoids help control the norm of the gradient of the loss function, and in turn mitigate some of the negative effects from clipping. In Figure~\ref{fig:tempered-sigmoids}, we visualize the influence of the scale $s$, inverse temperature $T$, and offset $o$ on the test performance of models trained with DP-SGD and tempered sigmoids $\phi_{s,T,o}$.

\begin{figure}[h]
\centering
\includegraphics[width=4.5cm]{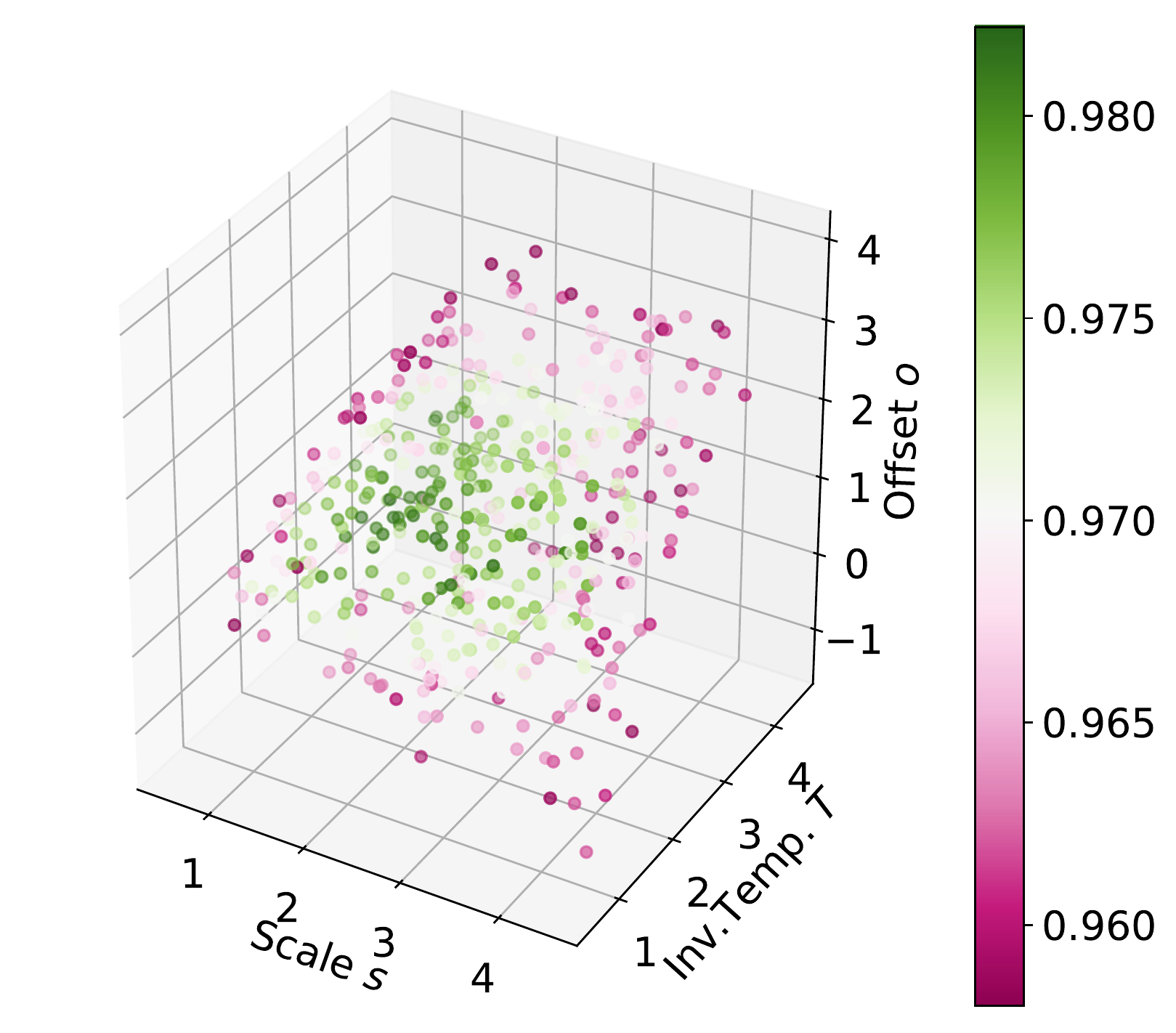}
\includegraphics[width=4.5cm]{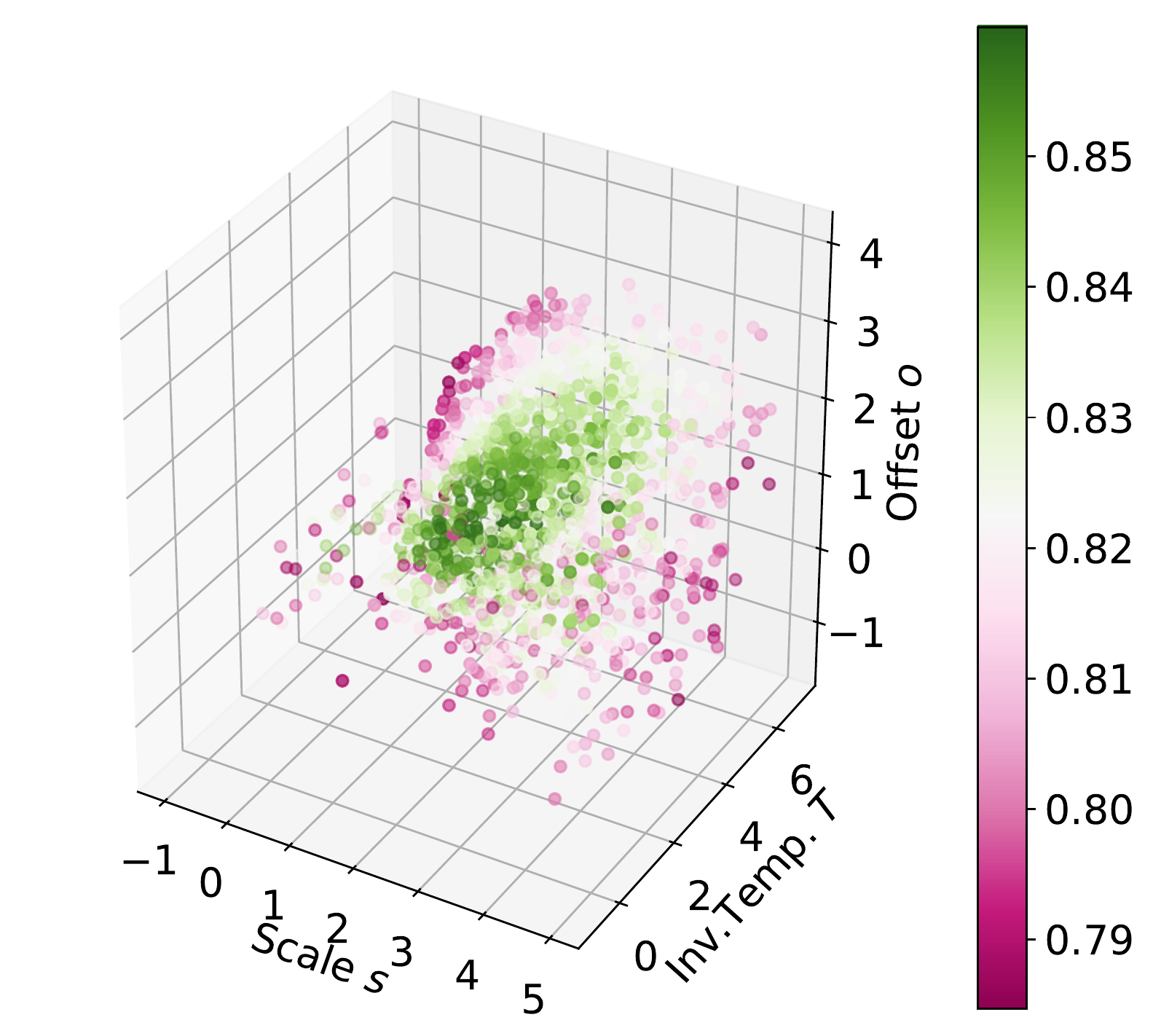}
\includegraphics[width=4.5cm]{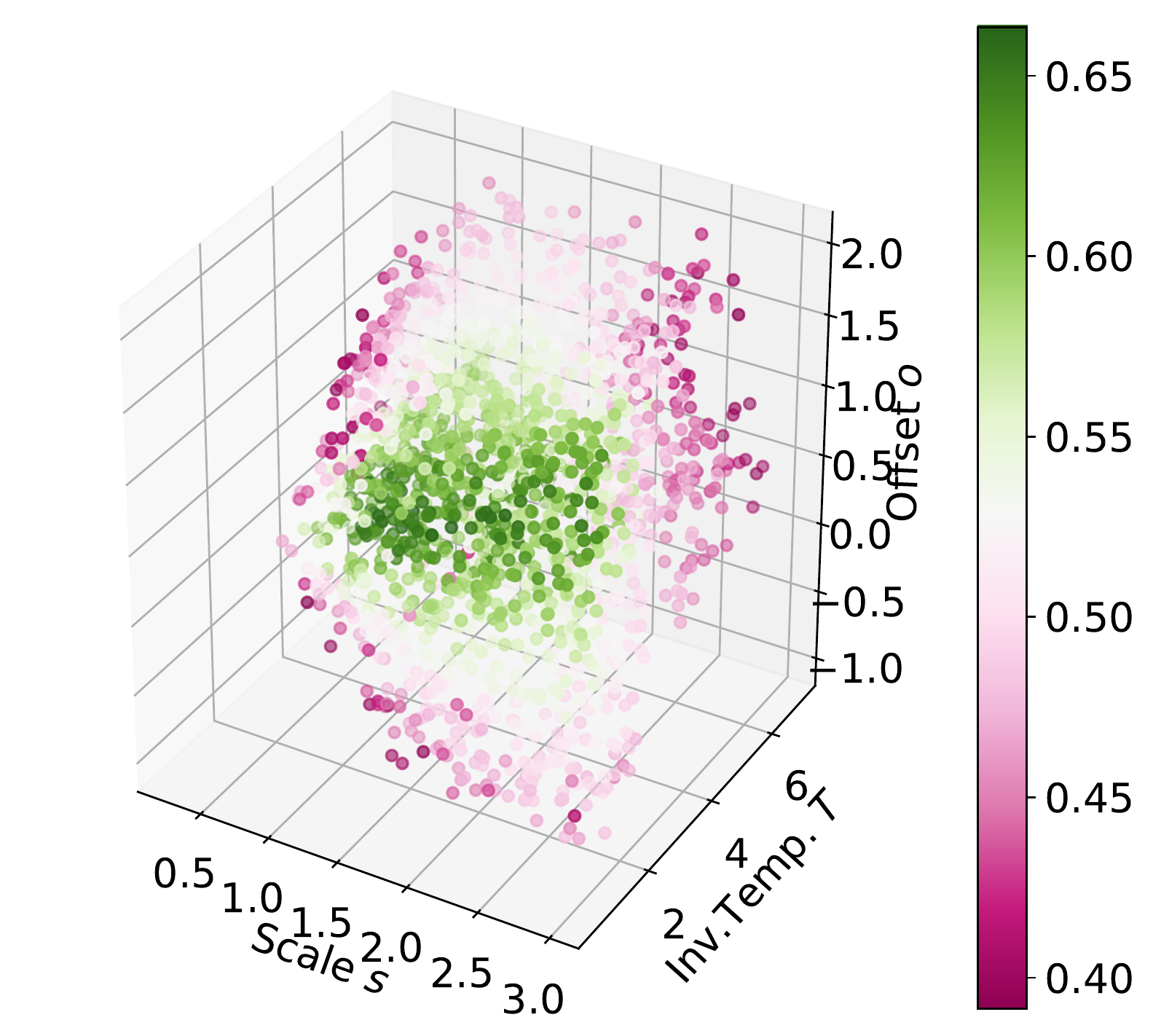}
\caption{Test accuracy of architectures with tempered sigmoids $\phi_{s,T,o}$ as a function the scale $s$, inverse temperature $T$, and offset $o$. Results are plotted for MNIST, FashionMNIST, and CIFAR10 from left to right. All models are trained with DP-SGD.}
\label{fig:tempered-sigmoids}
\end{figure}

Tempered sigmoids significantly outperform models trained with ReLU on all three datasets. On MNIST, the best tempered sigmoid achieves $98.1\%$ test accuracy whereas the baseline ReLU model trained to provide identical privacy guarantees ($\varepsilon=2.93$) achieved $96.6\%$ accuracy. This contributes to bridging the gap between privacy-preserving learning and non-private learning, which results in a test accuracy of $99.0\%$ for this architecture with both ReLU and \texttt{tanh}. On FashionMNIST, we achieve a best performing model of $86.0\%$ with tempered sigmoids in comparison with $81.9\%$ with ReLUs. A non-private model achieves $89.3\%$ with \texttt{tanh} and $89.4\%$ with ReLUs. On CIFAR10, the best  tempered sigmoid architectures achieve $66.0\%$ test accuracy whereas the ReLU variant obtained $61.6\%$ under the same privacy guarantees and the non-private baseline $76.6\%$.

From Figure~\ref{fig:tempered-sigmoids}, it appears clearly that a subset of tempered sigmoids performs best when learning with DP-SGD on the three datasets we considered. These form a cluster of points $(s, T, o)$ which result in models with significantly higher test accuracy. These points are colored in dark green. For each dataset, we compute the average value of the 10\% best-performing triplets $(s,T,o)$. On MNIST, the average triplet obtained is $(s,T,o)=(1.97,2.27,1.15)$,  on FashionMNIST $(s,T,o)=(2.27,2.61,1.28)$, and on CIFAR10 $(s,T,o)=(1.58, 3.00, 0.71)$. While this observation may not hold for other datasets, it is thus interesting to note here how these values happen to be close to the triplet setting $(s,T,o)=(2,2,1)$ for MNIST and FashionMNIST---and to a lesser extent for CIFAR10. Recall that this setting  corresponds exactly to the \texttt{tanh} function. 
For this reason, we explore  the particular case of \texttt{tanh} next. We seek to understand whether it is able to sustain the significant improvements of tempered sigmoids over ReLU for the datasets we considered.

\subsection{Improving the state-of-the-art on MNIST, FashionMNIST, and CIFAR10 with tanh}
\label{sec:tanh}

We now turn to the particular case of the \texttt{tanh} function to understand the broader implications of our results from Section~\ref{sec:family} for the three datasets considered: MNIST, FashionMNIST, and CIFAR10.
In the following experiment, we find that for these datasets positive results observed on the general family of tempered sigmoids can be reproduced with \texttt{tanh} alone, which is obtained by setting $s=2, T=2$, and $o=1$ in $\phi_{s,T,o}$. One of the  reasons we focus on the particular example of \texttt{tanh} is that changing the activation function to a \texttt{tanh} does not introduce new hyperparameters in learning: the values of $s,T,o$ need not be tuned if we choose to train architectures with a  \texttt{tanh}.

\begin{figure}[h]
\centering
\includegraphics[width=4.5cm]{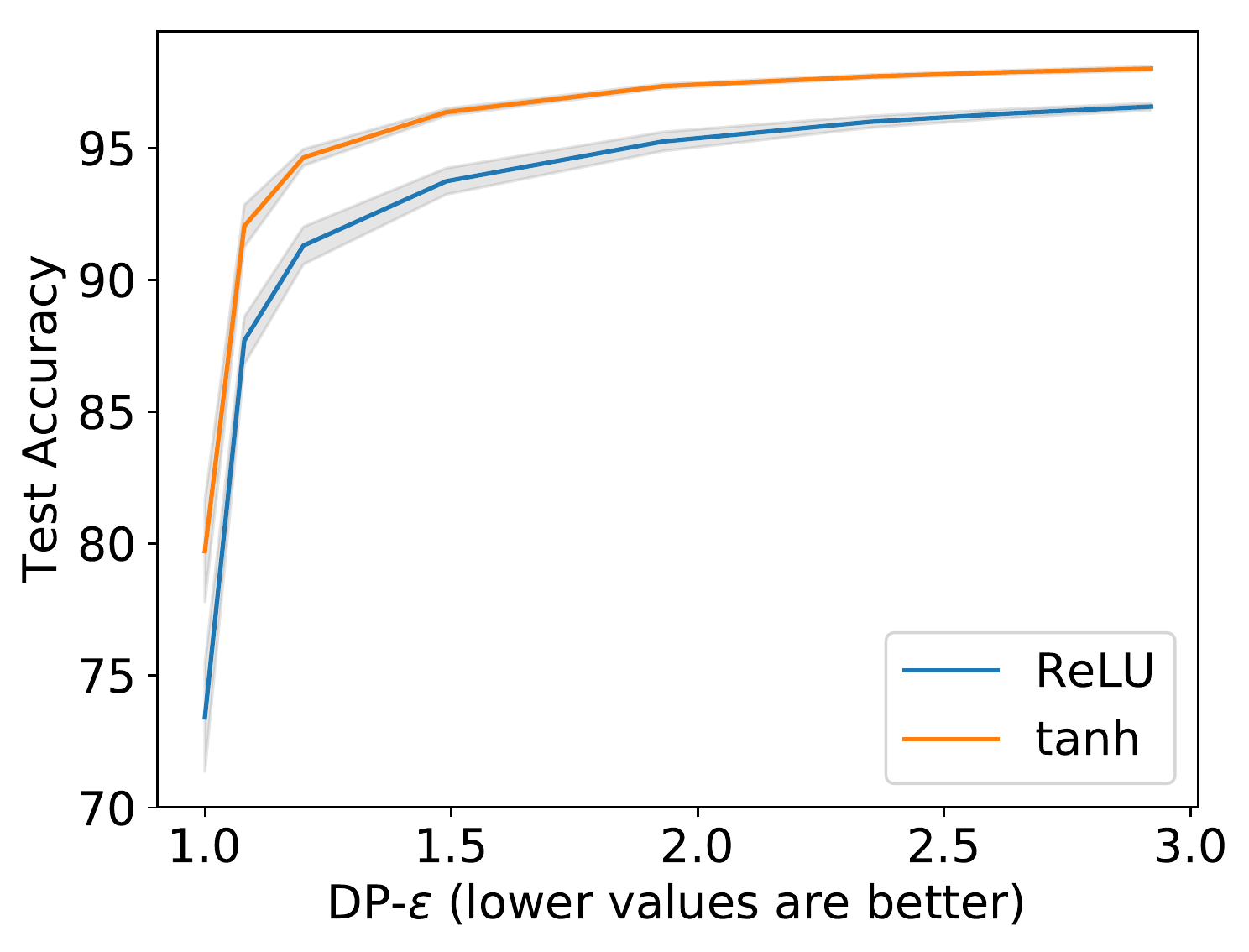}
\includegraphics[width=4.5cm]{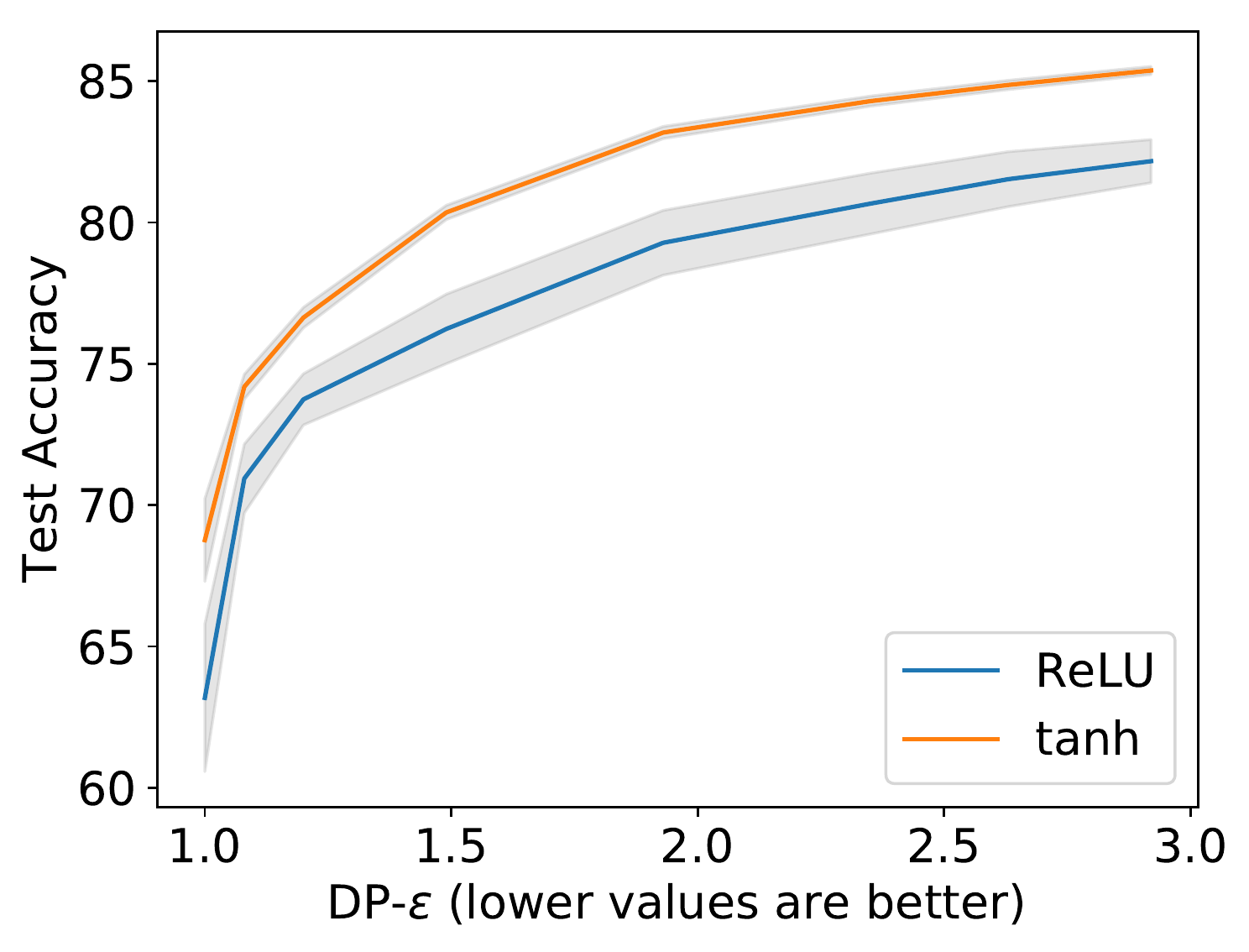}
\includegraphics[width=4.5cm]{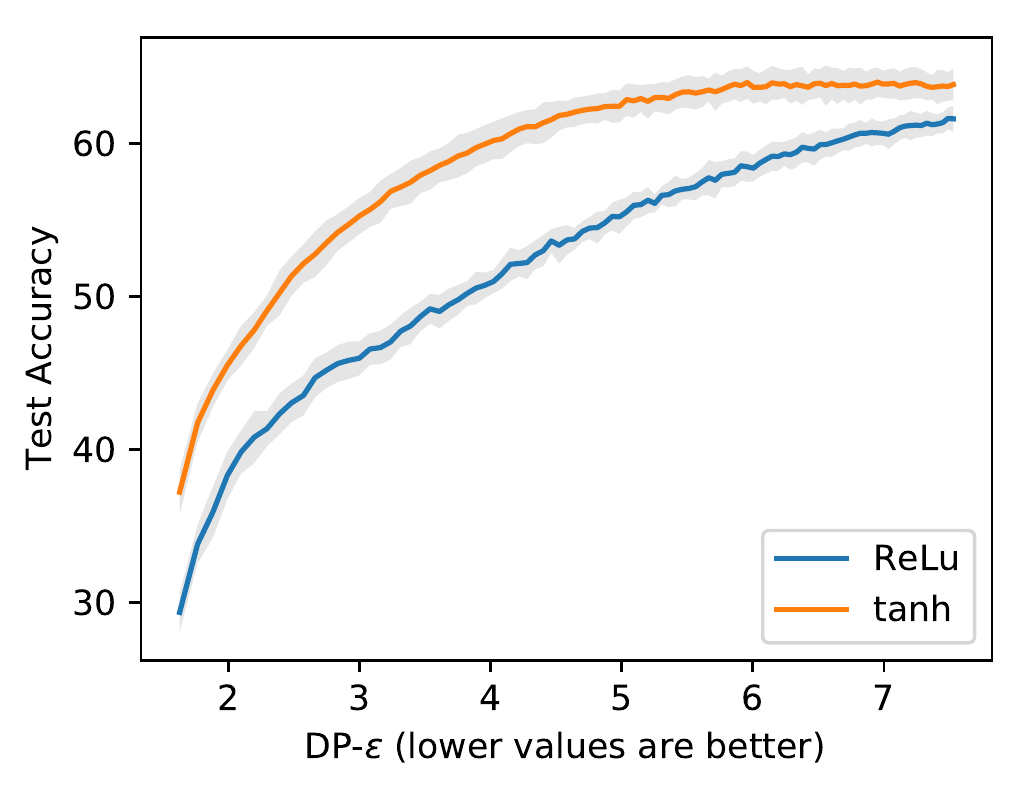}
\caption{Test accuracy as a function of the privacy loss when training a pair of models with DP-SGD on MNIST, FashionMNIST, and CIFAR10 (left to right). The only difference between the two models is the activation function for their hidden layer: ReLU or \texttt{tanh}. All other elements of the architecture (number, type, and dimension of layers) and the training algorithm (optimizer, learning rate, number of microbatches, clipping norm, and noise multiplier) are identical. Results  averaged over 10 runs.}
\label{fig:tanh-vs-relu-privacy-utility}
\end{figure}

\paragraph{Comparing performance.} The
\texttt{tanh} was an improvement on all three datasets and its performance is in line with the best test accuracy observed across tempered sigmoids on Figure~\ref{fig:tempered-sigmoids}. The test accuracy of the tanh model is $98.0\%$ on MNIST, $85.5\%$ on FashionMNIST, and $63.84\%$ on CIFAR10.
Figure~\ref{fig:tanh-vs-relu-privacy-utility} visualizes the  privacy-utility Pareto curve~\cite{avent2019automatic} of the ReLU and \texttt{tanh} models trained with DP-SGD for all three datasets. Rather than plotting the test accuracy as a function of the number of steps, we plot it as a function of the privacy loss $\varepsilon$ (but the privacy loss is a monotonically increasing function of the number of steps). The \texttt{tanh} models outperform their ReLU counterparts consistently regardless of the privacy loss $\varepsilon$ expended.

\begin{wrapfigure}{r}{6.1cm}
	\centering
	\includegraphics[width=6.1cm]{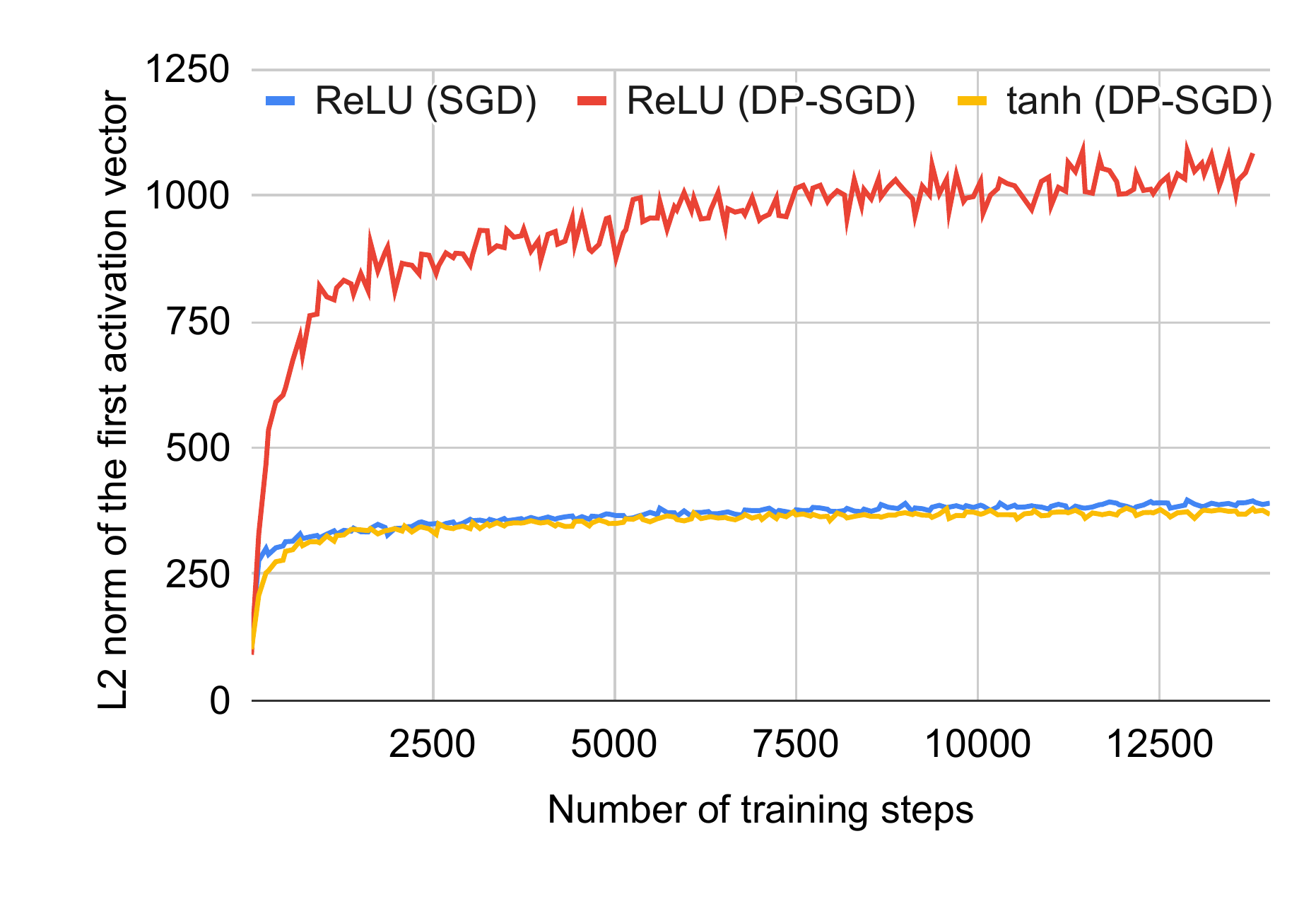}
	\caption{$\ell_2$ norm of the first conv activations. Three scenarios are plotted: (a) the model is trained without privacy using plain SGD, (b) the model is trained with ReLU activations with DP-SGD, and (c) the model is trained with tanh activations with DP-SGD.
	}
	\label{fig:tanh-vs-relu-activations}
\end{wrapfigure}

\paragraph{Impact of tanh on activation norms.} To explain why a simple change of activation functions has such
a large positive impact on the model's accuracy, we conjectured that the bounded nature of the \texttt{tanh}, and more generally tempered sigmoids, prevents activations from exploding during 
training. 

We monitored the $\ell_2$ norm of the
first layer's activations for our MNIST model while
it is being trained in three scenarios: (a)  without privacy using vanilla SGD and ReLU activations, 
(b)  with ReLU activations and DP-SGD, and (c) with \texttt{tanh} activations and DP-SGD. Activation  norms on  test data are visualized in Figure~\ref{fig:tanh-vs-relu-activations}. As conjectured in Section~\ref{sec:approach}, the activations of our ReLU model explode by a factor of $3$ when training with privacy compared to without privacy. Switching to  \texttt{tanh} brings down the norms of activations back to levels comparable with the activations of our non-private ReLU network. This helps us to learn with privacy---because it eliminates the negative effects of clipping and noising large gradients. By predisposing gradients to the operations performed by DP-SGD, less information is lost: the norm of unclipped gradients is closer to the clipping norm, and is also more adequate to the noise scale. We observe the same qualitative differences on other datasets, but do not repeat the plots due to space constraints.

\paragraph{Fine-tuning the optimizer.} To ensure that the comparison between ReLU and tempered sigmoids is fair, we now turn our attention to the training algorithm itself and verify that the superior behavior of \texttt{tanh} holds after a thorough hyperparameter search: this includes the number of filters $k$, learning rate, optimizer, batch size, and number of epochs. We find that it is  important to tailor algorithm and  hyperparameter choices to the specificities of private learning: an optimizer or learning rate that yields good results without privacy may not perform well with privacy. Among the hyperparameters mentioned above, it is  particularly important to fine-tune the learning rate to maximize performance given a fixed privacy budget. This is because the privacy budget limits the number of steps we can possibly take on the training set (as visualized on Figure~\ref{fig:tanh-vs-relu-privacy-utility}). For example, Table~\ref{tbl:batch-size} shows how  learning rates obtained through a hyperparameter search based on Batched Gaussian Process Bandits~\cite{desautels2014parallelizing} vary across the non-DP and DP settings when training on FashionMNIST, but that the choice of optimizer  (and in particular whether it is adaptive or not) does not influence results as much.

\begin{table}[h]
\centering
\begin{tabular}{|c|c|cc|cc|}
\multicolumn{2}{c}{~} &
\multicolumn{2}{c}{Non-private} & \multicolumn{2}{c}{Differentially-private}\\
    \multicolumn{1}{c}{Optimizer} & 
    \multicolumn{1}{c}{Epochs} & 
    \multicolumn{1}{c}{Learning Rate} & 
    \multicolumn{1}{c}{Test Accuracy} & 
    \multicolumn{1}{c}{Learning Rate} &
    \multicolumn{1}{c}{Test Accuracy}  \\
    \hline
    SGD & 40 & $1.07\cdot 10^{-1}$ & $90.3\%$ & $3.32\cdot 10^{-1}$ & $86.1\%$\\
    \hline
    Adam &  40  & $1.06\cdot 10^{-3}$ & $90.5\%$ & $1.32\cdot 10^{-3}$ & $86.0\%$\\
    \hline
\end{tabular}
\caption{Impact of learning rate on trade-off between accuracy and privacy. The privacy budget is fixed to $\varepsilon=2.7$ for all rows. 
A hyperparameter search is then conducted to find the best learning rate to train the model with or without differential privacy on FashionMNIST.}
\label{tbl:batch-size}
\end{table}

Table~\ref{tbl:summary} summarizes the results after performing this hyperparameter search for each of the three datasets considered in our experiments. We compare the non-private baseline and the DP-SGD with ReLU baseline to our DP-SGD approach with tempered sigmoids (instantiated by \texttt{tanh} here on our three datasets) after all hyperparameters have been jointly fined-tuned.
Even in their own individually-best setting, tempered sigmoids continue to consistently outperform ReLU with 98.1\% test accuracy (instead of 96.6\% for ReLU) on MNIST,  86.1\% test accuracy (instead of 81.9\% for ReLU) on FashionMNIST, 66.2\% test accuracy (instead of 61.6\% for ReLU) on CIFAR10.

\begin{table}[h]
    \centering
    \begin{tabular}{|c|c|c|c|c|}
    \multicolumn{1}{c}{Dataset} & 
    \multicolumn{1}{c}{Technique} & 
    \multicolumn{1}{c}{Acc.} & 
    \multicolumn{1}{c}{$\varepsilon$} &
    \multicolumn{1}{c}{$\delta$}
    \\
    \hline
    \hline
    \multirow{3}{*}{MNIST} & SGD w/ ReLU (not private) & 99.0\% & $\infty$ & 0 \\
    \cline{2-5}
     & DP-SGD w/ ReLU & 96.6\% & 2.93 & $10^{-5}$ \\
     & \textbf{DP-SGD w/ tempered sigmoid (tanh) [ours]} & \textbf{98.1\%} & \textbf{2.93} & $10^{-5}$ \\
    \hline
    \hline
    \multirow{3}{*}{FashionMNIST} & SGD w/ ReLU (not private) & 89.4\% & $\infty$ & 0 \\
    \cline{2-5}
     & DP-SGD w/ ReLU & 81.9\% & 2.7 & $10^{-5}$ \\
     & \textbf{DP-SGD w/ tempered sigmoid (tanh) [ours]} & \textbf{86.1\%} & \textbf{2.7} & $10^{-5}$ \\
    \hline
    \hline
    \multirow{3}{*}{CIFAR10} & SGD w/ ReLU (not private) & 76.6\% & $\infty$ & 0 \\
    \cline{2-5}
     & DP-SGD w/ ReLU                   & 61.6\%          & 7.53 & $10^{-5}$ \\
     & \textbf{DP-SGD w/ tempered sigmoid (tanh) [ours]}   & \textbf{66.2\%} & \textbf{7.53} & $10^{-5}$ \\
    \hline
    \end{tabular}
    \caption{Summary of results comparing ReLU to tempered sigmoids (represented here by the \texttt{tanh}) in their respective best performing setting (i.e., each row is the result of a hyperparameter search).
    }
    \label{tbl:summary}
\end{table}

\section{Conclusions}
\label{sec:conclusions}

Rather than first train a non-private model and later attempt to make it private, we bypass non-private training altogether and directly incorporate specificities of private learning in the selection of activation functions. Selecting a tempered sigmoid as the activation function renders the architecture more suitable for learning with differential privacy. This improves substantially upon the state-of-the-art privacy/accuracy trade-offs on three benchmarks which remain challenging for deep learning with differential privacy: MNIST, FashionMNIST, and CIFAR10. 
Future work may continue to explore this avenue: model architectures need to be chosen explicitly for privacy-preserving training
We expect that in addition to activation functions studied in our work, other architectural aspects can be modified to further reduce the observed gap in performance of private learning compared to non-private learning.
In addition, choosing the  parameters $(s,T,o)$ to be shared across all layers was not a necessity. We found that  layer-wise parameters did not improve results on our datasets, but it may be the case for different tasks:  this is related to the idea of setting layer-wise clipping norms~\cite{mcmahan2018general}.

\section*{Broader impact}

Our work helps make privacy-preserving training more practical. Our analysis and experimental results help practitioners make better  choices when design neural  architectures for privacy-preserving deep learning. In particular, the conclusions from our paper can readily be applied in real-world machine learning pipelines.  For this reason, we expect the broader impact of this work to be generally positive given the numerous applications of machine learning to sensitive datasets. This includes applications in domains like healthcare or language modeling.

\appendix

\end{document}